\documentclass{article}

    \PassOptionsToPackage{numbers, compress}{natbib}
\usepackage[preprint]{neurips_2024}

\usepackage[normalem]{ulem}

\usepackage[utf8]{inputenc} 
\usepackage[T1]{fontenc}    
\usepackage{hyperref}       
\usepackage{url}            
\usepackage{booktabs}       
\usepackage{amsfonts}       
\usepackage{nicefrac}       
\usepackage{microtype}      
\usepackage{xcolor}         
\usepackage{amsmath}
\usepackage{booktabs}
\usepackage{adjustbox}
\usepackage{multirow} 
\usepackage{colortbl} 
\usepackage{arydshln}

\title{VocalNet: Speech LLM with Multi-Token Prediction for Faster and High-Quality Generation}

\author{%
Yuhao Wang$^{1,2\ast}$ \quad Heyang Liu$^{1,2\ast}$ \quad Ziyang Cheng$^{3\ast}$ \quad Ronghua Wu$^2$ \quad Qunshan Gu$^{2}$ \\
\textbf{Yanfeng Wang}$^1$ \quad \textbf{Yu Wang}$^{1\dagger}$\\
$^1$Shanghai Jiao Tong University \quad $^2$ Ant Group \quad $^3$ Wuhan University\\
\texttt{\{colane, liuheyang, wangyanfeng622, yuwangsjtu\}@sjtu.edu.cn}\\
\texttt{\{r.wu, guqunshan.gqs\}@antgroup.com}\\
\texttt{icelookgoose@gmail.com}
}

\begin{document}

\maketitle

\renewcommand{\thefootnote}{\fnsymbol{footnote}}
\setcounter{footnote}{0}
\footnotetext{$^\ast$Equal contribution}
\footnotetext{$^\dagger$Corresponding author}
\renewcommand{\thefootnote}{\arabic{footnote}}

\begin{abstract}

Speech large language models (LLMs) have emerged as a prominent research focus in speech processing. We introduce VocalNet-1B and VocalNet-8B, a series of high-performance, low-latency speech LLMs enabled by a scalable and model-agnostic training framework designed for real-time voice interaction. Central to our contribution is the first application of multi-token prediction (MTP) to speech LLMs. This approach represents a paradigm shift from standard next-token prediction (NTP), offering simultaneous improvements in generation speed and quality. Informed by analysis of MTP's effect on speech generation and experimental comparisons, we designed a straightforward and highly effective MTP implementation. Experiments demonstrate that VocalNet performs on par with mainstream Omni LLMs even with limited training data, and significantly surpasses existing open-source speech LLMs. To foster reproducibility and community advancement, all model weights, inference code, training data, and framework implementations have been made publicly available at \url{https://github.com/SJTU-OmniAgent/VocalNet}.


\end{abstract}

\section{Introduction}

The development of speech interaction systems has shifted from traditional cascade-based architectures to end-to-end models. Traditional speech interaction systems typically adopt a cascade structure, consisting of automatic speech recognition (ASR), large language model (LLM), and text-to-speech (TTS) modules~\citep{shen2023hugginggpt, huang2024audiogpt, an2024funaudiollm}. However, this architecture often leads to system delays and information loss. {GPT-4o~\citep{gpt4o} demonstrates the potential of end-to-end speech interaction systems, namely speech LLMs, which process speech directly within a unified model.} This approach enhances the understanding and generation of speech content, and facilitates more natural audio interactions, improving real-time performance. As discussed in~\citet{chen2025minmo}, speech LLMs can be categorized into two types: native multimodal models and aligned multimodal models. Native multimodal models, such as Mini-Omni~\cite{xie2024mini}, Moshi~\cite{defossez2024moshi}, and GLM-4-Voice~\cite{zeng2024glm}, use a decoder-only Transformer to simultaneously decode both text and speech, achieving integration within a unified architecture. However, these models require large amounts of pretraining data and suffer from catastrophic forgetting. In contrast, aligned multimodal models, including LLaMA-Omni~\cite{fang2024llama}, Freeze-Omni~\cite{wang2024freeze}, and Qwen2.5-Omni~\cite{xu2025qwen}, incorporate separate speech encoders and decoders alongside an LLM backbone to handle speech understanding and generation. This approach better preserves the knowledge and reasoning capabilities of LLMs while requiring relatively less training data.



However, current research on aligned multimodal models has not yet deeply explored the modeling methods and training paradigms for speech generation.  Most existing models rely on autoregressive speech decoders that adopt the next-token prediction (NTP) paradigm for both training and inference. While this method has proven successful, it may not be the most efficient for speech modeling, given the complexity of speech signals. Compared to text, speech signals exhibit more intricate temporal characteristics and convey richer information. The length of a speech token sequence is often much longer than that of a corresponding text token sequence, leading to higher delays in the NTP process, which can be a challenge for real-time speech interactions.  Furthermore, individual speech tokens often lack distinct semantic meanings, as they represent very short time intervals. Human speech consists of structural elements, such as phonemes and syllables, which typically require multiple speech tokens to represent. The granularity mismatch between speech tokens and the underlying speech structure poses challenges for the NTP paradigm, which focuses on predicting only one token at a time. Inspired by recent advancements in LLMs~\citep{qi2020prophetnet, gloeckle2024better, cai2024medusa}, we investigate the potential of multi-token prediction (MTP) for speech LLMs. By analyzing the impact of MTP on speech generation, we identify limitations in previous implementations and propose an improved approach tailored to speech LLMs. Our findings show that, with limited training data, our MTP method not only accelerates the generation speed but also significantly improves speech quality.

Based on the proposed MTP implementation, we introduce \textbf{VocalNet-1B} and \textbf{VocalNet-8B}, speech interaction systems with high performance and low latency. Alongside the LLM backbone, VocalNet incorporates a speech encoder, an MTP decoder, and a vocoder. We also present a scalable, LLM-agnostic training framework that efficiently equips LLMs with real-time speech interaction capabilities. Experimental results show that VocalNet achieves performance comparable to advanced mainstream Omni LLMs like MiniCPM-o~\citep{MiniCPM-o-2.6} and Qwen2.5-Omni~\citep{xu2025qwen}, despite using much less training data, and significantly outperforms previous open-source speech LLMs like Freeze-Omni~\citep{wang2024freeze}. Moreover, while previous work has only released model weights and inference code, the data processing pipelines and training frameworks often remain opaque, which has hindered further research. To foster further academic exploration of speech LLMs and encourage broader community participation, we would open-source our model training code, inference code, model weights, and the data used in this work, providing valuable resources for the academic community. In summary, our contributions can be summarized as follows:

\begin{itemize}

\item We propose a scalable, model-agnostic training framework to cost-effectively enable LLMs with real-time voice interaction capabilities, advancing the development of speech LLMs.





\item We introduce the MTP approach for speech LLMs and propose an effective MTP implementation. Through detailed analysis and experimental comparison, we identify the limitations of previous method, and further propose a simple and more efficient MTP implementation specifically for speech LLMs. This approach not only accelerates speech generation but also archives consistent quality improvements, providing a new insight for speech LLMs.

\item We conduct extensive experiments that demonstrate the superior voice interaction performance of VocalNet with a limited training corpus, highlighting the efficiency, scalability, and cost-effectiveness of the proposed framework and the effectiveness of the MTP approach.


\end{itemize}

\section{Related Work}

\subsection{End-to-End Speech Interaction System}

End-to-end speech interaction systems have become a key research focus in the speech processing community. As discussed in~\citet{chen2025minmo}, speech LLMs can be categorized into two types: native multimodal models and aligned multimodal models. Native multimodal speech LLMs generate tokens for both modalities using a unified backbone. These models can be further divided into two categories: one type, represented by Mini-Omni~\cite{xie2024mini}, Moshi~\cite{defossez2024moshi}, {PSLM\citep{mitsui2024pslm}} and SLAM-Omni~\cite{chen2024slam}, adopts a multi-stream architecture that simultaneously generates audio and text outputs. The other type, including {OmniFlatten~\citep{zhang2024omniflatten}, }GLM-4-Voice~\cite{zeng2024glm}{, SpiRit LM~\citep{nguyen2025spirit}} and Baichuan-Omni-1.5~\cite{li2025baichuan}, generates interleaved audio and text outputs to handle both modalities. However, these models require large amounts of speech-text pairs for training to avoid catastrophic forgetting. Even using a large amount of training data, their knowledge and reasoning capabilities often fall short compared to similar-sized LLMs.

Alternatively, aligned multimodal models introduce separate encoders, decoders, and vocoders for speech processing. This architecture has the advantage of preserving the original abilities of LLMs while also generating high-quality speech responses. LLaMA-Omni~\cite{fang2024llama} uses a non-autoregressive method based on connectionist temporal classification (CTC)~\citep{graves2006connectionist} for speech generation. Although it offers low latency, the quality of the generated speech is relatively poor. Freeze-Omni~\cite{wang2024freeze}, MiniCPM-o~\cite{MiniCPM-o-2.6}, MinMo~\cite{chen2025minmo} and VITA-1.5~\cite{fu2025vita} all employ autoregressive speech decoders trained with the next-token prediction task for speech generation. Qwen2.5-Omni~\cite{xu2025qwen} introduces a dual-track autoregressive Transformer decoder architecture for speech decoding, which enables more natural streaming inference without modifying the training process. However, the superiority of this dual-stream framework in speech modeling still requires further investigation in future research.

\subsection{Multi-token Prediction}
Multi-token prediction has emerged as an important advancement in language modeling, offering improvements in sample efficiency, reasoning capabilities, and inference speed. 
The concept of multi-token prediction was initially explored by~\citet{qi2020prophetnet}, who proposed training models to predict several future tokens in parallel.  Building upon this foundation, ~\citet{gloeckle2024better} introduced a refined architecture that incorporated multiple output heads operating over a shared model backbone. Their approach demonstrated that multi-token prediction could lead to models that are both better and faster. Furthermore, ~\citet{cai2024medusa} proposed a speculative decoding method based on multi-token prediction to accelerate LLM inference.

In the context of speech generation, several works have employed group modeling techniques to implement multi-token prediction. SLAM-Omni~\citep{chen2024slam} proposes a semantic group modeling approach to accelerate speech token generation and model training. This method partitions the speech token sequence into fixed-size groups and uses a linear layer to reconstruct each group embedding into multiple speech tokens. Similarly, IntrinsicVoice~\citep{zhang2024intrinsicvoice} introduces GroupFormer, a non-autoregressive Transformer module to perform token reconstruction. While group modeling methods can accelerate speech generation, they often lead to quality degradation, particularly as the group size increases.

\section{VocalNet}

\subsection{Model Architecture}

\begin{figure*}[t]
  \centering
  \includegraphics[width=0.9\textwidth]{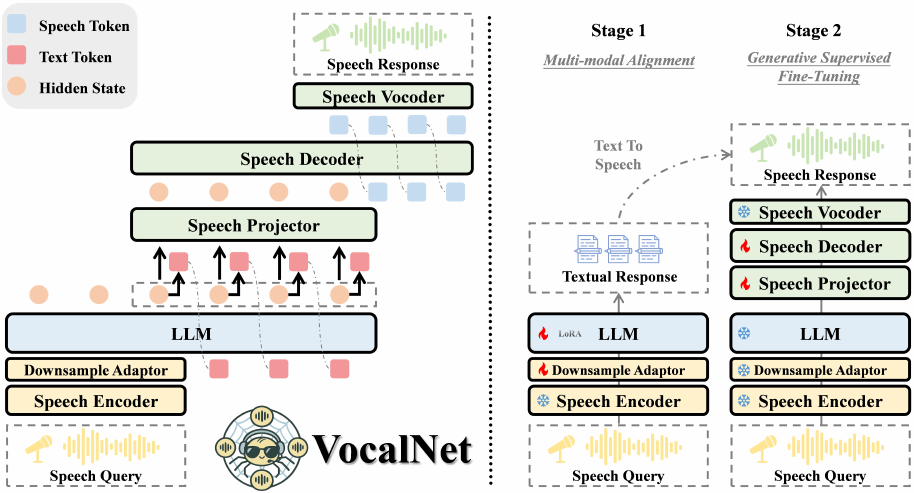}
  \caption{On the \textbf{left}: The architecture of the VocalNet model. On the \textbf{right}: A depiction of VocalNet's dual-stage training strategy.}
  \label{vocalNet}
\end{figure*}

The model architecture of VocalNet is illustrated in Figure~\ref{vocalNet}. Align with prior work, VocalNet consists of a speech encoder to convert waves into speech representations, a pre-trained LLM backbone and a speech decoder for speech token generation. A downsample adaptor is added after the speech encoder to achieve a lower frame rate, and a speech projector to bridge the dimension gap between the LLM hidden state and decoder inputs. The generated speech token is sent to the speech vocoder, in which the corresponding speech response is constructed. This architecture effectively preserves the capabilities inherent in the pre-trained LLM, thus significantly reducing the data requirement for training compared with native multimodal models. In the following statement, $\boldsymbol{x}^s$ refers to the raw speech query, $\boldsymbol{y}^t$ represents the generated text response and $\boldsymbol{y}^s$ stands for the speech response.

\paragraph{Speech Query Encoding}


The speech encoder $E$ processes the raw input speech query $\boldsymbol{x}^s$ to produce a high-level representation $z$ with length $l$: $\boldsymbol{z} = E(\boldsymbol{x}^s) = (z_0, z_1, ..., z_l)$, which encapsulates rich semantic information. After that, the downsample adaptor transforms the speech feature $\boldsymbol{z}$ into semantic-condensed embedding with a lower frame rate. Through a concatenation-based projection module, it reduces the sequence length by a factor of $k$, yielding $\boldsymbol{z}'$, and applies linear transformations with ReLU activation to generate $\boldsymbol{z}_o$, which will be fed into the LLM backbone, as expressed in
\begin{equation}
\begin{aligned}
\boldsymbol{\boldsymbol{z}'_i} &= \text{Concate}(z_{ir}, z_{ir+1}, ..., z_{(i+1)r-1})\\
\boldsymbol{z}_o &= W_2 (\text{ReLU}(W_1\boldsymbol{z'}+b_1)) + \boldsymbol{b}_2
\end{aligned}
\end{equation}


where $W_1$ and $W_2$ are weight matrices, $\boldsymbol{b}_1$ and $\boldsymbol{b}_2$ are bias vectors. This process ensures semantic preservation and alignment with the LLM’s feature space.


\paragraph{LLM}
The LLM functions as the core module, processing the compressed representation $\boldsymbol{z}_o$ to extract linguistic and contextual information, yielding hidden states $\boldsymbol{h}_{LLM}$. These states enable the generation of the corresponding textual response $\boldsymbol{y}^t$ and are essential in speech generation.

\paragraph{Speech Response Generation}
The speech decoder need to model both the LLM hidden states $\boldsymbol{h}_{LLM}$ and the speech embedding simultaneously, but the spaces represented by these two are typically different~\citep{wang2024freeze}. To address this space gap, we introduce a speech projector that transforms $\boldsymbol{h}_{LLM}$ into $\boldsymbol{v}_{LLM}$. The speech decoder then utilizes these vectors to autoregressively generate a sequence of discrete speech tokens $\boldsymbol{s}$. Finally, a pre-trained speech vocoder, incorporating a chunk-aware flow matching model derived from \citep{du2024cosyvoice} along with HifiGAN~\citep{kong2020hifi}, constructs the mel-spectrogram from the speech tokens $\boldsymbol{s}$ and then synthesizes the corresponding speech waveform response $\boldsymbol{y}^s$.

\subsection{Training Strategy}  


We adopt a dual-stage training strategy as shown in the right part of Figure~\ref{vocalNet}: Multi-Modal Alignment and Generative Supervised Fine-Tuning, as categorized in~\cite{ji2024wavchat}. In the {first stage}, VocalNet is trained using speech queries and text responses $(\boldsymbol{x}^s \xrightarrow{} \boldsymbol{y}^t)$. The speech encoder is frozen to maintain its capability of extracting meaningful speech representations, while the downsample adaptor is unfrozen to facilitate the alignment between speech and text features.  The LLM backbone is trained using LoRA to strengthen its multi-modal performance while keep its original capabilities like general knowledge and reasoning. In this stage, we compute the cross-entropy loss on text tokens which helps the model learn to understand speech inputs. In the second stage, VocalNet is trained using speech query and speech response $(\boldsymbol{x}^s \xrightarrow{} \boldsymbol{y}^s)$. During this stage, the major components of the model are frozen, and the speech projector and speech decoder are trained to generate high-quality speech tokens $s$ corresponding to the ground-truth speech response $y^s$. In this stage, we compute the cross-entropy loss on speech tokens to guide the model in generating accurate speech responses.

Our staged training approach decomposes the task into two manageable steps, allowing for a more stable and controlled training process. While our framework could support training both speech understanding and generation within a single stage, our initial experiments did not reveal significant advantages to this approach. In contrast, the two-stage method offers greater stability and control.

\subsection{Streaming Speech Decoding}
\label{sec:Streaming Speech Decoding}

To enable efficient speech decoding in streaming scenarios while ensuring high-quality non-streaming speech decoding, we employ two attention mask mechanisms tailored for complete sequence processing and real-time speech generation respectively, inspired by \cite{MiniCPM-o-2.6}. During the generative supervised fine-tuning stage, these two mask mechanisms are used simultaneously in a batch, allowing the model to flexibly adapt to diverse decoding requirements.


\begin{figure*}[t]
  \centering
  \includegraphics[width=0.95\textwidth]{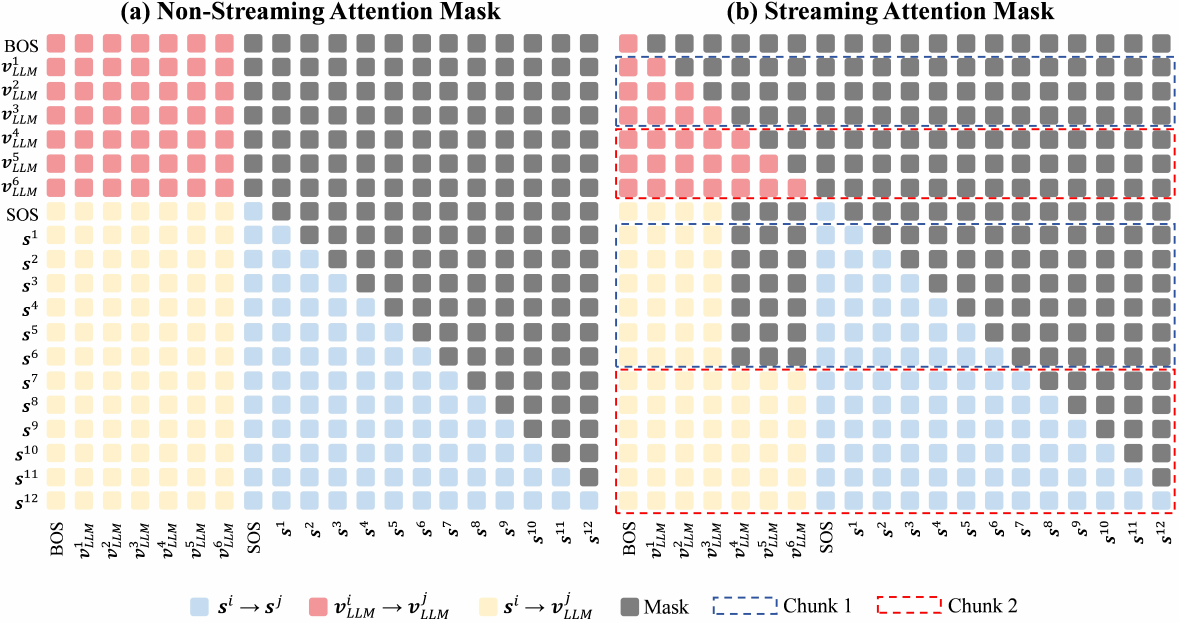}
  \caption{(a) Non-Streaming Attention Mask: $\boldsymbol{v}_{LLM}^i$ attends to itself and all text positions, and $\boldsymbol{s}^i$ attends to itself, all text positions, and its previous speech positions; (b) Streaming Attention Mask: $\boldsymbol{v}_{LLM}^i$ attends to itself and its previous text positions, and $\boldsymbol{s}^i$ attends to itself, chunk-limited text positions, and its previous speech positions.}
  \label{fig:Attention Mask}
\end{figure*}


\paragraph{Non-Streaming Attention Mask}

The non-streaming attention mask as shown in Figure~\ref{fig:Attention Mask} (a), is optimized for scenarios involving the one-time processing of complete input sequences. BOS and SOS refer to `begin of stream' and `switch of stream', two identified special tokens. The yellow blocks refer to the attended text positions during speech generation, and the blue and red ones are the attended positions within the same modality. {In this mode, the text hidden states $\boldsymbol{v}_{LLM}$ generated by the speech projector from $\boldsymbol{h}_{LLM}$ are fully visible to themselves, while the attention for the speech component adheres to an autoregressive property, meaning each speech token $\boldsymbol{s}^i$ depends solely on itself and preceding tokens. Additionally, speech tokens $\boldsymbol{s}^i$ have unrestricted access to the text hidden states $\boldsymbol{v}_{LLM}$, leveraging global contextual information comprehensively.}


{Given the text hidden state $\boldsymbol{v}_{LLM} \in \mathbb{R}^{L_t}$ with length $L_t$ and the speech hidden state $\boldsymbol{s} \in \mathbb{R}^{L_s}$ with length $L_s$, the attention mask $\boldsymbol{A} \in \{0, 1\}^{(L_t + L_s) \times (L_t + L_s)}$ for a single instance is defined:}




\begin{equation}
	A_{i,j} = \begin{cases}
	      1 &  i \leq L_{t} \\
	      1 &  i > L_{t}, i\geq j \\
            0 &  \text{otherwise}
		   \end{cases}
\end{equation}


\paragraph{Streaming Attention Mask}

The streaming attention mask as shown in Figure~\ref{fig:Attention Mask} (b), is specifically designed for real-time speech generation, supporting the incremental processing of input sequences. {In this mode, both the text hidden states $\boldsymbol{v}_{LLM}$ and speech hidden states $\boldsymbol{s}$ are constrained by an autoregressive mask, permitting access only to preceding positions.}


Let the speech sequence length $L_s$ be divided into chunks of length $C_s$, with each along with increased visible real text positions (excluding BOS token) of length $C_t$. In Figure~\ref{fig:Attention Mask} (b), $C_s$ and $C_t$ is shown as 6 and 3 respectively. The streaming mask is formally defined as follows:



\begin{equation}
	A_{i,j} = \begin{cases}
	      1 &  i \leq L_{t}, i\geq j  \\
	      1 &  i > L_{t}, i\geq j > L_{t} \\
            1 &  i > L_{t}, j \leq min(L_{t}, \lceil (i - L_t-1) / C_s \rceil \cdot C_t + 1) \\
            0 &  \text{otherwise}
		   \end{cases}
\end{equation}



\section{Multi-Token Prediction for Speech Generation}
\subsection{Motivation}
Many previous works have employed the next-token prediction~(NTP) task to train speech decoder~\citep{fang2024llama, wang2024freeze}, using an autoregressive~(AR) model that predicts one token at each inference step. However, a significant frequency disparity exists between text tokens (\textasciitilde3Hz)~\citep{li2025fast, defossez2024moshi} and speech tokens (\textasciitilde25Hz)~\citep{du2024cosyvoice}, which results in speech sequences being much longer than their corresponding textual format. This inherent characteristic of speech presents a critical challenge, as the single-token prediction mechanism leads to extended speech generation times. This limitation becomes particularly critical in real-time voice interaction systems, where low-latency generation is essential.


Additionally, human speech exhibits a complex hierarchical structure, comprising elements such as phonemes, syllables, prosody, and semantic features. Unlike text tokens, which often carry explicit semantic meaning, speech tokens generally lack such clarity on their own, as they correspond to very short, low-level acoustic segments (e.g., each speech token in CosyVoice 2 represents approximately 40 ms of audio). Consequently, multiple speech tokens are typically required to represent a single phoneme or semantic unit. This mismatch between the granularity of speech tokens and the underlying speech structures we aim to model presents a challenge for NTP paradigm, which focuses solely on predicting one token at a time. Under limited data conditions, the model may struggle to learn such intricate structural complexity of speech effectively, potentially leading to suboptimal performance in capturing the full richness of spoken language.


Inspired by recent advancements in LLMs~\citep{gloeckle2024better,li2024eagle, cai2024medusa}, we introduce the multi-token prediction (MTP) approach to address the above challenges and improve speech generation efficiency. In this section, we will first explore the potential impact of MTP in speech modeling, and then provide a detailed discussion of its implementation and the design of the model architecture.

 \subsection{Analysis of the Impact of MTP in Speech Generation}
\label{sec:AnalysisMTPImpact}

\subsubsection{Mitigating Error Accumulation}
\label{sec:Error Accumulation}

\begin{figure}[t]
  \centering
  \includegraphics[width=0.8\textwidth]{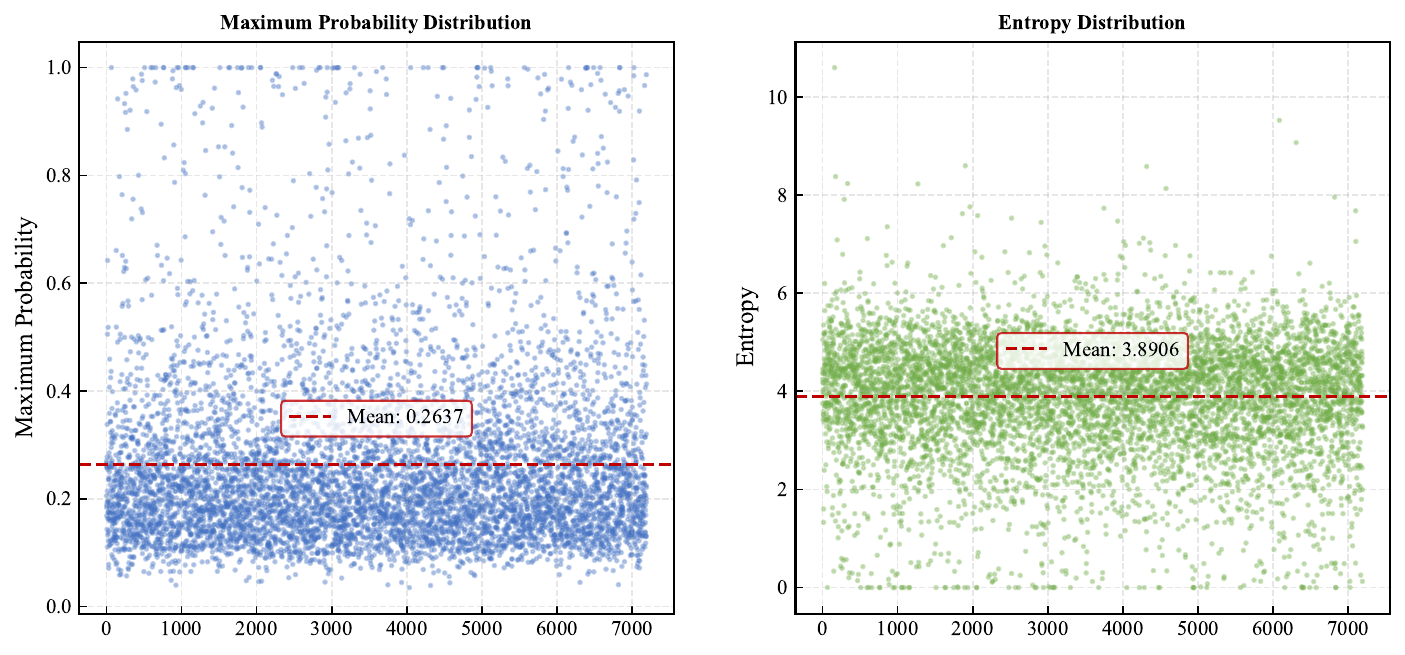}
  \caption{Distribution of maximum probabilities and entropy values for 70k predicted speech tokens from VocalNet-1B, trained with the NTP task. Red dashed lines represent the means.}
  \label{fig:scatter_distributions}
\end{figure}



Autoregressive models are commonly trained using teacher forcing, where the model is provided with the correct {history} tokens as input during training. However, during inference,  the model generates outputs {based on the predicted history in the autoregressive manner}, which leads to the accumulation of errors. In speech generation tasks, we observe that the multinomial distributions predicted by our model tend to exhibit a flattened pattern. Figure~\ref{fig:scatter_distributions} illustrates the distribution of maximum probabilities and entropy values across 70k predicted speech token distributions from VocalNet-1B trained with the NTP task. The results show that the maximum probabilities predominantly cluster below 0.25, while the entropy values generally exceed 3. Our observation indicates that most of the speech predictions contain multiple tokens with similar probabilities, reflecting high uncertainty in the model’s predictions. This phenomenon contributes to the worsening of error accumulation during speech generation. {With an MTP loss added to the model training, this issue could be mitigated. The MTP loss is expressed as follows:}


\begin{equation}
\begin{aligned}
    \mathcal{L}_{\text{MTP}} &= -\sum_{\boldsymbol{x}} \log q(\boldsymbol{x}_{t+1:t+K} | \boldsymbol{x}_{\leq t}),\\
    &=-\sum_{\boldsymbol{x}}\sum_{k} \log q(\boldsymbol{x}_{t+k} | \boldsymbol{x}_{\leq t}),\\
\end{aligned}
\label{eq:mtploss}
\end{equation}
where $q$ denotes the model's predictions, $t$ represents the current time step, $\boldsymbol{x}$ refers to the data sample, $\boldsymbol{x}_{\leq t}$ denotes the historical sequence up to time $t$, and $K > 1$ indicates the number of future steps that need to be predicted.

As shown in Equation~\ref{eq:mtploss}, the MTP loss function compels the model to learn to generate the correct future tokens \(\boldsymbol{x}_{t+k}\) based on incomplete history \(\boldsymbol{x}_{<t}\). This strategy allows the model to better handle the inherent uncertainty in the autoregressive process, leading to more accurate and robust predictions even when faced with noisy input history. As a result, the model becomes less dependent on perfect target sequences and more resilient to the noise introduced during inference.

\subsubsection{Effectively Capturing Local Patterns in Speech}

The MTP loss, by directly learning the joint distribution \( p(\boldsymbol{x}_{t+1:t+k} | \boldsymbol{x}_{<t}) \) of speech tokens, encourages the model to capture short-term temporal relationships and understand the underlying local dependencies within speech. In practice, multiple MTP modules can generate predictions for several future tokens based on the hidden state of the final layer of the speech decoder. This setup enables the model to anticipate the potential impact of future tokens while predicting the current token, effectively modeling local dependencies between them.

From an information-theoretic perspective, \citet{gloeckle2024better} demonstrates that in a two-token prediction scenario, the MTP loss emphasizes the relative mutual information $I_{p \parallel q}(X; Y)$, where $X$ and $Y$ are consecutive tokens. By minimizing this term, the model can better leverage the mutual information between adjacent tokens under the true distribution $p$, improving its ability to predict tokens while capturing their subtle interconnections. This is crucial for speech modeling, as it helps the model understand the local patterns inherent in speech.

Local patterns are particularly important in speech modeling. Neighboring speech tokens typically correspond to related units, such as phonemes or syllables. Understanding these relationships is vital for maintaining coherence and rhythm in speech. By encouraging the model to capture these local dependencies, the MTP loss enhances its ability to generate speech that is not only contextually accurate but also naturally fluent. In this way, the MTP loss plays a crucial role in helping the model learn short-term dependencies, enabling it to more effectively handle the complex structures that characterize natural speech.

\subsection{Implementation of MTP}
\label{sec:Implementation of MTP}

\paragraph{Group Modeling Method}

\begin{figure}[t]
  \centering
  \includegraphics[width=1\textwidth]{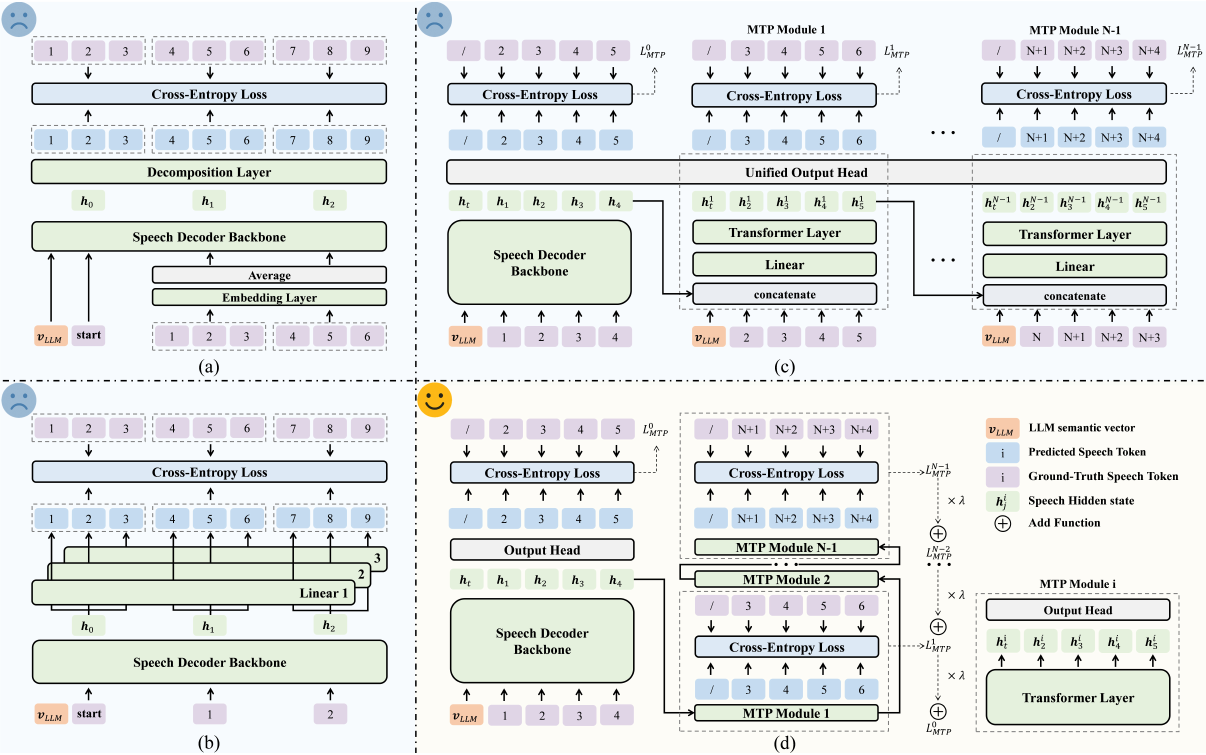}
  \caption{Illustration of various accelerate implementations. (a): Group Modeling; (b): MTP-Parallel-Linear; (c): MTP-DeepSeek; (d): Our MTP implementation.}
  \label{fig:MTP}
\end{figure}

To accelerate speech token generation, previous works have adopted the Group Modeling method \citep{chen2024slam, zhang2024intrinsicvoice} to enable multi-token prediction, as shown in Figure~\ref{fig:MTP}(a). This approach partitions the speech token sequence into fixed-size groups and merges all tokens within each group into a single embedding. After processing these merged embeddings through the backbone network, a decomposition layer reconstructs each group embedding into multiple individual speech tokens. Specifically, SLAM-Omni \citep{chen2024slam} employs a simple linear layer for decomposition, whereas IntrinsicVoice \citep{zhang2024intrinsicvoice} utilizes a non-autoregressive Transformer module with multiple learnable queries. However, these methods typically degrade speech quality due to inevitable information loss caused by tokens merging, as well as the disruption of temporal dependencies within each group. Furthermore, the fixed-size group limits the ability to dynamically adjust the acceleration ratio during inference.

\paragraph{MTP Implementation in LLMs}
Inspired by the implementation of MTP in ~\citet{gloeckle2024better} and DeepSeek-V3~\citep{liu2024deepseek}, we designed two speech decoder architectures to achieve multi-token prediction, namely MTP-Parallel-Linear and MTP-DeepSeek. As shown in Figure~\ref{fig:MTP}(b), MTP-Parallel-Linear parallelly predicts $n$ additional tokens using independent linear output heads, a method widely used in LLMs due to its simplicity and efficiency. However, speech is a continuous physical signal, and maintaining temporal dependencies between tokens is crucial. While this architecture generates tokens in parallel, it fails to explicitly model these dependencies, which can result in less coherent and natural speech, especially as the number of output heads increases.

On the other hand, as shown in Figure~\ref{fig:MTP}(c), MTP-DeepSeek generates tokens sequentially, preserving the causal chain for token prediction at each depth. However, during training, this implementation inputs the ground truth $x_{i+k}$ to the $k$-th MTP module to predict $x_{i+k+1}$ and computes the loss of a teacher-forced next-step prediction. Consequently, this implementation actually optimizes the loss function $-\sum_{\boldsymbol{x}} \sum_{k} \log q(\boldsymbol{x}_{t+k} | \boldsymbol{x}_{\leq t+k-1})$, which is essentially the same as the NTP loss. As a result, while this approach enables multi-token prediction, it does not effectively alleviate error accumulation or help capture local patterns in speech as discussed in section~\ref{sec:AnalysisMTPImpact}.

\paragraph{Our MTP Implementation}

Building on the strengths and limitations of the aforementioned MTP approaches, we propose a simple yet more effective MTP implementation tailored for speech LLMs. Since speech is a continuous signal that relies on temporal dependencies and contextual coherence between tokens, our approach, as shown in Figure~\ref{fig:MTP}(d), utilizes $N-1$ sequential Transformer layers as MTP modules. This design enables the prediction of \(N\) speech tokens in a single inference step while preserving the temporal relationships between these tokens. To fully leverage the two key advantages of MTP discussed in Section~\ref{sec:AnalysisMTPImpact}, unlike MTP-DeepSeek, we use the previous hidden states of the MTP module rather than ground truth tokens as input.

In detail, let $\boldsymbol{h}_{1:(L_t+t )}^0$ denote the hidden state generated by the speech decoder backbone, with input $v_{LLM}$ and $t$ speech tokens. This state is sequentially processed through $N-1$ MTP modules:
\begin{equation}
\boldsymbol{h}_{1:(L_t+t)}^k = MTP_k(\boldsymbol{h}_{1:(L_t+t)}^{k-1})
\end{equation}
where $\boldsymbol{h}_{1:(L_t+t)}^k$ represents the hidden state output of the $k$-th MTP module, with $k \in \{1, 2, \dots, N-1\}$. This layer-wise propagation preserves the causal dependencies of the speech sequence. The resulting $N$ hidden states at index $L_t+t$, $\boldsymbol{h}_{L_t+t}^0, \boldsymbol{h}_{L_t+t}^1, \dots, \boldsymbol{h}_{L_t+t}^{N-1}$, are then fed into $N$ independent output heads to produce token predictions:
\begin{equation}
p_{t+k+1}^k = OutHead_k(\boldsymbol{h}_{L_t+t}^k) = Linear_k(RMSNorm(\boldsymbol{h}_{L_t+t}^k))
\end{equation}
where $k \in \{0, 1, \dots, N-1\}$, and $p_{t+k+1}^k$ denotes the predicted probability distribution for the $(t+k+1)$-th token.


The objective of our MTP implementation is to minimize the prediction error at each depth of the MTP modules, which is computed by averaging the cross-entropy loss across the outputs of all \(N-1\) MTP modules and the speech decoder. Each module's contribution to the loss is weighted by a factor \(\lambda^k\), where \(k\) corresponds to the depth of the MTP module (ranging from 0 to \(N-1\)). More formally, the loss function is given by:
\begin{equation}
\begin{aligned}
\mathcal{L}_{MTP} = \sum_{k=0}^{N-1} \lambda^k \, CrossEntropy(p_{k+1:L_s}^k, s_{k+1:L_s})
\end{aligned}
\end{equation}
where $L_s$ is the total length of the speech token sequence, and $s_{k+1:L_s}$ denotes the ground-truth tokens from index $k+1$ to $L_s$. Here, the decay factor $\lambda \in (0,1)$ controls the relative importance of predictions at different depths of the MTP modules. Specifically, the decay factor \(\lambda\) assigns higher weights \(\lambda^k\) to the losses from earlier layers (smaller \(k\)), as these layers typically produce more reliable and immediate predictions. Conversely, losses from deeper layers (larger \(k\)), which tend to have higher uncertainty, receive progressively lower weights due to the exponential decay of \(\lambda^k\). This design enables the model to prioritize accuracy in the short-term predictions, while still benefiting from deeper-layer predictions that capture broader temporal context.

\section{Experiments Setup}

\subsection{Datasets}
The training corpus used for VocalNet includes VoiceAssistant-400K from Mini-Omni and UltraChat from SLAM-Omni~\cite{xie2024mini, chen2024slam}. VoiceAssistant-400K contains about 470K entries specifically generated by GPT-4o, providing query audios and response transcriptions. We obtain a cleaned version by removing instances with over-long responses, resulting in a modified set of 430K query-response pairs. For UltraChat, we decompose multi-round conversations into multiple single rounds, for the initial rounds of many dialogues are not provided and the context is typically uncorrelated. The processed UltraChat consists of around 300K entries. The response speech tokens for the aforementioned datasets are generated with CosyVoice2-0.5B~\cite{du2024cosyvoice}. In total, the VocalNet training set consists of 732K examples, with a total duration of approximately 6,000 hours—significantly less than other advanced open-source models, such as Baichuan-Omni-1.5 (887K hours in multi-modal pretraining) and Minmo (approximately 1.4M hours of audio data).





\subsection{Model Configuration}
\label{sec:model_config}
We propose VocalNet-1B and VocalNet-8B built upon LLaMA-3.2-1B-Instruct~\footnote{\url{https://huggingface.co/meta-llama/Llama-3.2-1B-Instruct}} and LLaMA-3.1-8B-Instruct~\footnote{\url{https://huggingface.co/meta-llama/Llama-3.1-8B-Instruct}} respectively. Both models utilize Whisper-large-v3~\cite{radford2023robust} as the speech encoder and the flow-matching model along with the HiFi-GAN vocoder from CosyVoice 2 to construct the speech response. The downsample adaptor is a 2-layer linear for feature compression with a downsample factor of 5. The speech projector consists of 2-layer Llama decoder layers. The speech decoder comprises 4-layer Llama decoder layers with 2048 hidden size, 32 attention heads, and an 8192-dimensional feed-forward network. Each MTP module is constructed using a single-layer Llama decoder layer with a linear output head. For streaming decoding, the chunk size \( C_s \) and \( C_t \) are set to 15 and 5 respectively.


\subsection{Training and Evaluation Details}

The training of VocalNet is carried out in two distinct phases. In the first phase, we focus on training the downsample adaptor and the LLM. The second phase targets the speech projector and the speech decoder. For both phases, the learning rate is $2 \times 10^{-4}$, and a cosine annealing learning rate schedule is applied, with a warmup ratio of 0.03. All training processes are performed on A100 GPUs.



To evaluate the capabilities of voice interaction, we utilize the English subsets from OpenAudioBench~\citep{li2025baichuan}, which include AlpacaEval~\citep{alpaca_eval}, Llama Questions~\citep{nachmani2023spoken}, TriviaQA~\citep{joshi2017triviaqa}, Web Questions~\citep{berant2013semantic}. For the evaluation process, we employ Qwen-max~\footnote{\url{https://qwenlm.github.io/blog/qwen2.5-max/}} to score and determine the correctness of responses. Following Baichuan-omni-1.5~\citep{li2025baichuan}, the score for Llama Questions is calculated as the percentage of answers deemed correct. For Web Questions and TriviaQA, we scale the scores and normalize them to a range of 0 to 10. For AlpacaEval, the score range is set to 1 to 10.

Furthermore, we employ two metrics to evaluate the quality of the generated speech. To assess the overall speech quality, we use the UTMOS~\citep{saeki2022utmos} to predict mean opinion scores (MOS). For evaluating the alignment between speech and text responses, we transcribe the speech by Whisper-large-v3~\cite{radford2023robust} and calculate the word error rate (WER), regarding the recognition results as the hypothesis and corresponding text response as the transcription. 


\section{Experiments Results}
\subsection{Overall Result}
\label{sec: overallresult}

\begin{table}[t]
  \caption{Comparison with different speech LLMs and omni LLMs on OpenAudioBench. \textbf{Bold} indicates the optimal result in each subgroup and \underline{underline} indicates the suboptimal result.}
  \label{tab: result1_main}
  \centering
  \resizebox{1\textwidth}{!}{
  \begin{tabular}{cc>{\centering\arraybackslash}c>{\centering\arraybackslash}p{1.5cm}>{\centering\arraybackslash}p{2.5cm}>{\centering\arraybackslash}p{1.5cm}>{\centering\arraybackslash}p{2.5cm}}
  \toprule
    \toprule
{\textbf{Model}} & {\textbf{LLM size}} & {\textbf{Modality}} & {\textbf{AlpacaEval}} & {\textbf{Llama Questions}} & {\textbf{TriviaQA}} & {\textbf{Web Questions}} \\ 
\hline
\hline
\multirow{2}{*}{Mini-Omni} & \multirow{2}{*}{0.5B} & s$\rightarrow$t & 1.84 & 2.7 & 0.12 & 0.22\\
& & s$\rightarrow$s & 1.80 & 2.7 & 0.08 & 0.20\\
\arrayrulecolor[gray]{0.5}
\cdashline{2-7}
\arrayrulecolor{black} 
\multirow{2}{*}{SLAM-Omni}& \multirow{2}{*}{0.5B} & s$\rightarrow$t & 3.50 & 29.4 & 0.39 & 0.84 \\
& & s$\rightarrow$s & 3.01 & 26.7 & 0.34 & 0.69\\
\arrayrulecolor[gray]{0.5}
\cdashline{2-7}
\arrayrulecolor{black} 
\multirow{2}{*}{VocalNet-1B (VA)}& \multirow{2}{*}{1B}  & \cellcolor[rgb]{ .902,  .902,  .902}{s$\rightarrow$t} & \cellcolor[rgb]{ .902,  .902,  .902} 5.38 & \cellcolor[rgb]{ .902,  .902,  .902} 70.3 & \cellcolor[rgb]{ .902,  .902,  .902} 3.38 & \cellcolor[rgb]{ .902,  .902,  .902} 4.93\\
&  & \cellcolor[rgb]{ .902,  .902,  .902}{s$\rightarrow$s} & \cellcolor[rgb]{ .902,  .902,  .902} 4.83 & \cellcolor[rgb]{ .902,  .902,  .902} 61.0 & \cellcolor[rgb]{ .902,  .902,  .902} 2.78 & \cellcolor[rgb]{ .902,  .902,  .902} 4.47\\
\multirow{2}{*}{VocalNet-1B}& \multirow{2}{*}{1B}  & \cellcolor[rgb]{ .902,  .902,  .902}{s$\rightarrow$t} & \cellcolor[rgb]{ .902,  .902,  .902} \textbf{5.79} & \cellcolor[rgb]{ .902,  .902,  .902} \textbf{71.7} & \cellcolor[rgb]{ .902,  .902,  .902} \textbf{3.60} & \cellcolor[rgb]{ .902,  .902,  .902} \textbf{5.16}\\
&  & \cellcolor[rgb]{ .902,  .902,  .902}{s$\rightarrow$s} & \cellcolor[rgb]{ .902,  .902,  .902} \textbf{5.03} & \cellcolor[rgb]{ .902,  .902,  .902} \textbf{63.7} & \cellcolor[rgb]{ .902,  .902,  .902} \textbf{3.06} & \cellcolor[rgb]{ .902,  .902,  .902} \textbf{4.68}\\
\hline
\multirow{2}{*}{LLaMA-Omni}& \multirow{2}{*}{8B} & s$\rightarrow$t & 5.31 & 69.7 & 4.44 & 5.44\\
& & s$\rightarrow$s & 3.89 & 55.1 & 2.44 & 4.00 \\
\arrayrulecolor[gray]{0.5}
\cdashline{2-7}
\arrayrulecolor{black} 
\multirow{2}{*}{Freeze-Omni}& \multirow{2}{*}{7B} & s$\rightarrow$t & 4.51 & 77.7 & 5.32 & 6.41\\
& & s$\rightarrow$s & 2.99 & 60.2 & 3.53 & 4.78\\
\arrayrulecolor[gray]{0.5}
\cdashline{2-7}
\arrayrulecolor{black} 
\multirow{2}{*}{GLM-4-Voice} & \multirow{2}{*}{9B} & s$\rightarrow$t & 5.86 & 77.4 & 4.95 & 5.56\\
& & s$\rightarrow$s & 5.27 & 64.3 & 4.63 & 5.40\\
\arrayrulecolor[gray]{0.5}
\cdashline{2-7}
\arrayrulecolor{black} 
\multirow{2}{*}{Baichuan-Omni-1.5} & \multirow{2}{*}{7B} & s$\rightarrow$t & 5.20 & 77.6 & 5.72 & 6.12\\
& & s$\rightarrow$s & 4.10 & 61.2 & 4.13 & 5.18\\
\arrayrulecolor[gray]{0.5}
\cdashline{2-7}
\arrayrulecolor{black} 
\multirow{2}{*}{MiniCPM-o} & \multirow{2}{*}{8B} & s$\rightarrow$t & 6.13 & 77.2 & \textbf{6.43} & \textbf{7.16}\\
& & s$\rightarrow$s & 4.95 & 65.8 & 4.99 & \underline{6.22}\\
\arrayrulecolor[gray]{0.5}
\cdashline{2-7}
\arrayrulecolor{black} 
\multirow{2}{*}{Minmo*} & \multirow{2}{*}{8B} & s$\rightarrow$t & {-} & 78.9 & 4.83 & 5.50\\
& & s$\rightarrow$s & \textbf{6.48} & 64.1 & 3.75 & 3.99\\
\arrayrulecolor[gray]{0.5}
\cdashline{2-7}
\arrayrulecolor{black} 
\multirow{2}{*}{Qwen2.5-Omni} & \multirow{2}{*}{8B} & s$\rightarrow$t & 6.01 & \underline{79.0} & 5.89 & \underline{6.88}\\
& & s$\rightarrow$s & 5.73 & \textbf{76.3} & \underline{5.59} & \textbf{6.70}\\
\arrayrulecolor[gray]{0.5}
\cdashline{2-7}
\arrayrulecolor{black} 
\multirow{2}{*}{VocalNet-8B (VA)}& \multirow{2}{*}{8B} & \cellcolor[rgb]{ .902,  .902,  .902}{s$\rightarrow$t} & \cellcolor[rgb]{ .902,  .902,  .902} \underline{7.05} & \cellcolor[rgb]{ .902,  .902,  .902} 77.1 & \cellcolor[rgb]{ .902,  .902,  .902} 6.15 & \cellcolor[rgb]{ .902,  .902,  .902} 6.34\\
& & \cellcolor[rgb]{ .902,  .902,  .902}{s$\rightarrow$s} & \cellcolor[rgb]{ .902,  .902,  .902} 6.30 & \cellcolor[rgb]{ .902,  .902,  .902} 71.4 & \cellcolor[rgb]{ .902,  .902,  .902} 5.24 & \cellcolor[rgb]{ .902,  .902,  .902} 5.81\\
\multirow{2}{*}{VocalNet-8B}& \multirow{2}{*}{8B} & \cellcolor[rgb]{ .902,  .902,  .902}{s$\rightarrow$t} & \cellcolor[rgb]{ .902,  .902,  .902} \textbf{7.12} & \cellcolor[rgb]{ .902,  .902,  .902} \textbf{79.5} & \cellcolor[rgb]{ .902,  .902,  .902} \underline{6.24} & \cellcolor[rgb]{ .902,  .902,  .902} 6.48\\
& & \cellcolor[rgb]{ .902,  .902,  .902}{s$\rightarrow$s} & \cellcolor[rgb]{ .902,  .902,  .902} \underline{6.37} & \cellcolor[rgb]{ .902,  .902,  .902} \underline{73.1} & \cellcolor[rgb]{ .902,  .902,  .902} \textbf{5.67} & \cellcolor[rgb]{ .902,  .902,  .902} 6.16\\
\hline
\hline
  \end{tabular}
  }
\end{table}

Table~\ref{tab: result1_main} presents the performance of VocalNet in voice assistant scenario compared to other mainstream speech LLMs and omni LLMs that possess speech interaction abilities. All models are inferred in a speech-to-speech (s2s) setting with the default parameters. For $s \rightarrow t$ modality, the text response is assessed, while for $s \rightarrow s$, the speech response is transcribed by Whisper-large-v3 and then evaluated. The result for Minmo is taken from its paper, as its model has not been released. {For both sizes of VocalNet, we propose the evaluation of two versions, where VocalNet (VA) is trained with only VoiceAssistant-400K and the other uses the combination of VoiceAssitant-400K and UltraChat.}


For tiny speech LLMs (LLM size $\leq$ 1B), VocalNet-1B substantially outperforms Mimi-Omni and SLAM-Omni, both developed based on Qwen2-0.5B.  Even though our model size is around twice as compared to these models, we achieve significant gains (i.e. 71.7\% accuracy for text response on LLaMA Questions compared to 2.7\% and 29.4\%). It is even more gratifying that VocalNet-1B has performance advantages on specific datasets compared to some base-sized speech LLMs ($\sim$8B). On AlpacaEval, it achieves better scores compared to LLaMA-Omni, Freeze-Omni, and Baichuan-Omni-1.5. On LLaMA Questions, it surpasses LLaMA-Omni. In addition, VocalNet-1B preserves the potential of further improvement by just extending its training datasets, because the performance across these four datasets has been enhanced when adding UltraChat alongside VoiceAssistant-400K.

For base-sized speech LLMs, VocalNet-8B achieves performance comparable to MiniCPM-o and Qwen2.5-Omni, and steadily outperforms the other models. On AlpacaEval, LLaMA Questions, and TriviaQA, VocalNet-8B ranks among the top-2 models and achieves three first-place finishes, demonstrating its superior overall performance among the evaluated models. For Web Questions, VocalNet ranks third, slightly behind MiniCPM-o and Qwen2.5-Omni. 


To quantify the multi-modal response alignment and the acoustic quality, we also present the results for WER and UTMOS. As shown in Table~\ref{tab: result2_wer_utmos}, VocalNet-1B surpasses other tiny models across all metrics. By utilizing additional training data, VocalNet-1B exhibits gain on multi-modal alignment with consistent acoustic score. VocalNet-8B maintains its strength in acoustic quality, and achieves the second-lowest WER, surpassed only by Qwen2.5-Omni. 

\begin{table}
  \caption{Comparison with different models in response alignment and acoustic performance. Bold indicates the optimal result in each subgroup and \underline{underline} indicates the suboptimal result.}
  \label{tab: result2_wer_utmos}
  \centering
  \resizebox{1\textwidth}{!}{
  \begin{tabular}{ccccccccccc}
  \toprule
  \toprule
  \multirow{2}{*}{\textbf{Model}} & \multicolumn{2}{c}{\textbf{AlpacaEval}} & \multicolumn{2}{c}{\textbf{Llama Questions}} & \multicolumn{2}{c}{\textbf{TriviaQA}} & \multicolumn{2}{c}{\textbf{Web Questions}} & \multicolumn{2}{c}{\textbf{Avg}} \\
  & \textbf{\small WER} & \textbf{\small UTMOS} & \textbf{\small WER} & \textbf{\small UTMOS} & \textbf{\small WER} & \textbf{\small UTMOS} & \textbf{\small WER} & \textbf{\small UTMOS} & \textbf{\small WER} & \textbf{\small UTMOS} \\
  \hline\hline
  \textbf{Mini-Omni} & 20.78 & 4.429 & 5.20 & 4.428 & 7.43 & 4.428 & 8.51 & 4.433 & 8.66 & 4.430 \\
  \arrayrulecolor[gray]{0.5}
  \cdashline{2-11}
  \arrayrulecolor{black} 
  \textbf{SLAM-Omni} & 5.52 & 4.439 & 5.55 & 4.467 & 6.16 & 4.470 & 6.50 & 4.461 & 6.17 & 4.464 \\
  \arrayrulecolor[gray]{0.5}
  \cdashline{2-11}
  \arrayrulecolor{black} 
  \textbf{VocalNet-1B (VA)} & \cellcolor[rgb]{.902,.902,.902}\textbf{3.43} & \cellcolor[rgb]{.902,.902,.902}\textbf{4.495} & \cellcolor[rgb]{.902,.902,.902}\underline{3.65} & \cellcolor[rgb]{.902,.902,.902}\textbf{4.498} & \cellcolor[rgb]{.902,.902,.902}\textbf{5.97} & \cellcolor[rgb]{.902,.902,.902}\textbf{4.499} & \cellcolor[rgb]{.902,.902,.902}\underline{6.40} & \cellcolor[rgb]{.902,.902,.902}\underline{4.489} & \cellcolor[rgb]{.902,.902,.902}\underline{5.66} & \cellcolor[rgb]{.902,.902,.902}\textbf{4.495} \\
  \arrayrulecolor[gray]{0.5}
  \cdashline{2-11}
  \arrayrulecolor{black} 
  \textbf{VocalNet-1B} & \cellcolor[rgb]{.902,.902,.902}\textbf{3.43} & \cellcolor[rgb]{.902,.902,.902}\underline{4.491} & \cellcolor[rgb]{.902,.902,.902}\textbf{3.27} & \cellcolor[rgb]{.902,.902,.902}\underline{4.497} & \cellcolor[rgb]{.902,.902,.902}\underline{6.73} & \cellcolor[rgb]{.902,.902,.902}\underline{4.486} & \cellcolor[rgb]{.902,.902,.902}\textbf{4.88} & \cellcolor[rgb]{.902,.902,.902}\textbf{4.493}  & \cellcolor[rgb]{.902,.902,.902}\textbf{5.31} & \cellcolor[rgb]{.902,.902,.902}\underline{4.491} \\
  \hline
  \textbf{LLaMA-Omni} & 6.00 & 3.942 & 10.00 & 4.003 & 20.93 & 3.965 & 14.60 & 3.935 & 15.90 & 3.956 \\
  \arrayrulecolor[gray]{0.5}
  \cdashline{2-11}
  \arrayrulecolor{black} 
  \textbf{Freeze-Omni} & 14.33 & {4.377} & 14.20 & {4.417} & 20.39 & {4.404} & 18.25 & {4.398} & 18.31 & {4.401} \\
  \arrayrulecolor[gray]{0.5}
  \cdashline{2-11}
  \arrayrulecolor{black} 
  \textbf{GLM-4-Voice} & 18.71 & 4.025 & 14.45 & 4.152 & 8.33 & 4.306 & 6.08 & 4.214 & 8.99 & 4.228 \\
  \arrayrulecolor[gray]{0.5}
  \cdashline{2-11}
  \arrayrulecolor{black} 
  \textbf{Baichuan-omni-1.5} & 20.84 & 4.082 & 22.82 & 4.332 & 22.36 & 4.401 & 23.29 & 4.350 & 22.67 & 4.347 \\
  \arrayrulecolor[gray]{0.5}
  \cdashline{2-11}
  \arrayrulecolor{black} 
  \textbf{MiniCPM-o} & 15.35 & 4.102 & 5.73 & 4.228 & 8.08 & 4.128 & 8.94 & 4.125 & 8.72 & 4.137 \\
  \arrayrulecolor[gray]{0.5}
  \cdashline{2-11}
  \arrayrulecolor{black} 
  \textbf{Qwen2.5-Omni} & \textbf{2.41} & 4.299 & \textbf{0.93} & 4.315 & \textbf{1.13} & 4.339 & {4.68} & 4.363 & \textbf{2.63} & 4.342 \\
   \arrayrulecolor[gray]{0.5}
  \cdashline{2-11}
  \arrayrulecolor{black} 
  \textbf{VocalNet-8B (VA)} & \cellcolor[rgb]{.902,.902,.902}\underline{2.65} & \cellcolor[rgb]{.902,.902,.902}\textbf{4.490} & \cellcolor[rgb]{.902,.902,.902} {3.00} & \cellcolor[rgb]{.902,.902,.902}\textbf{4.503} & \cellcolor[rgb]{.902,.902,.902}{5.02} & \cellcolor[rgb]{.902,.902,.902}\textbf{4.499} & \cellcolor[rgb]{.902,.902,.902}\underline{4.21} & \cellcolor[rgb]{.902,.902,.902}\underline{4.485} & \cellcolor[rgb]{.902,.902,.902}{4.26} & \cellcolor[rgb]{.902,.902,.902}\textbf{4.493} \\
   \arrayrulecolor[gray]{0.5}
  \cdashline{2-11}
  \arrayrulecolor{black} 
  \textbf{VocalNet-8B} & \cellcolor[rgb]{.902,.902,.902}{4.71} & \cellcolor[rgb]{.902,.902,.902}\underline{4.489} & \cellcolor[rgb]{.902,.902,.902}\underline{2.68} & \cellcolor[rgb]{.902,.902,.902}\underline{4.500} & \cellcolor[rgb]{.902,.902,.902}\underline{4.04} & \cellcolor[rgb]{.902,.902,.902}\underline{4.482} & \cellcolor[rgb]{.902,.902,.902}\textbf{3.11} & \cellcolor[rgb]{.902,.902,.902}\textbf{4.492} & \cellcolor[rgb]{.902,.902,.902}\underline{3.56} & \cellcolor[rgb]{.902,.902,.902}\underline{4.489} \\
  \hline\hline
  \end{tabular}
  }
\end{table}

\subsection{MTP Implementation}
\paragraph{MTP Implementation Method.}
In this section, we conduct experiments with the five MTP implementations discussed in Section~\ref{sec:Implementation of MTP}, utilizing the LLaMA-3.2-1B as the LLM backbone and trained with the VoiceAssistant-400K dataset. Results are shown in Table~\ref{tab: result3_mtp}. Group-linear and Group-Trans denote the group modeling approaches employed in SLAM-omni and IntrinsicVoice respectively. We test the group sizes of 3 and 5. The results show that while group modeling can improve the generation speed of speech tokens, it leads to a decline compared to NTP. This is especially noticeable with a larger group size, where both metrics exhibit considerable deterioration.



\begin{table}
  \caption{Comparison with different Implementation of MTP. Bold
indicates the optimal result.}
  \label{tab: result3_mtp}
  \centering
  \begin{tabular}{lcccc}
  \toprule
  \toprule
{\textbf{Method}}  & {\textbf{Group Size/Module Num}} & {\textbf{Speedup Ratio}}
& \textbf{WER$\downarrow$} & \textbf{UTMOS$\uparrow$}   \\ 
\hline
\hline
Baseline(NTP)& - & $1\times$ & {10.62} & 4.488 \\
\arrayrulecolor[gray]{0.5}
\hline
\arrayrulecolor{black} 
\multirow{2}{*}{Group-Linear} & 3 & $3\times$ & 11.50 & 4.488  \\
 & 5 & $5\times$ & 17.61  & 4.414  \\
\arrayrulecolor[gray]{0.5}
\cdashline{2-5}
\arrayrulecolor{black} 
\multirow{2}{*}{Group-Trans} & 3 & $3\times$ & 14.34 & {4.489} \\
& 5 & $5\times$ & 17.90  & 4.468  \\

\arrayrulecolor[gray]{0.5}
\hline
\arrayrulecolor{black} 
\multirow{3}{*}{MTP-Parallel-Linear} & \multirow{3}{*}{5} & $1\times$ & 8.61 & 4.492  \\
& & $3\times$ & 8.00  & {4.494}  \\
& & $5\times$ & 10.57  & 4.467  \\
\arrayrulecolor[gray]{0.5}
\cdashline{2-5}
\arrayrulecolor{black} 
\multirow{3}{*}{MTP-DeepSeek} & \multirow{3}{*}{5} & $1\times$ & 9.14  & 4.493  \\
& & $3\times$ & 9.02  & \textbf{4.498}  \\
& & $5\times$ & 18.23  & 4.488  \\
\arrayrulecolor[gray]{0.5}
\hline
\arrayrulecolor{black} 
\multirow{3}{*}{MTP-VocalNet} & \multirow{3}{*}{5} & $1\times$ & 6.84  &  {4.494}  \\
& & $3\times$ & \textbf{5.66}  & 4.495  \\
& & $5\times$ & 6.46  & 4.486  \\
\hline
\hline
  \end{tabular}
\end{table}

\begin{table}
  \caption{Comparison with different numbers of MTP modules utilized in the training and inferring phase. Bold indicates the optimal result and \underline{underline} indicates the suboptimal result.}
  \label{tab: result4_mpt_num}
  \centering
  \small
  \setlength{\tabcolsep}{5pt}
   \resizebox{1\textwidth}{!}{
  \begin{tabular}{cccccccccccccc}
  \toprule
    \toprule
\multirow{2}{*}{\textbf{Module Num}} & \multirow{2}{*}{\textbf{Speedup}} & \multicolumn{2}{c}{\textbf{AlpacaEval}} & \multicolumn{2}{c}{\textbf{Llama Questions}} & \multicolumn{2}{c}{\textbf{TriviaQA}} & \multicolumn{2}{c}{\textbf{Web Questions}}  & \multicolumn{2}{c}{\textbf{Avg}}  \\ 
 &  & \textbf{\small WER} & \textbf{\small UTMOS} & \textbf{\small WER} & \textbf{\small UTMOS} & \textbf{\small WER} & \textbf{\small UTMOS} & \textbf{\small WER} & \textbf{\small UTMOS}& \textbf{\small WER} & \textbf{\small UTMOS} \\ 
\hline
\hline
\multirow{2}{*}{3} & $1\times$ & 5.38 & 4.489 & 5.24 & \textbf{4.504} & 7.59 & \textbf{4.500} & 9.23 & 4.484 & 7.79 & 4.493\\
\arrayrulecolor[gray]{0.5}
\cdashline{2-12}
\arrayrulecolor{black} 
 & $3\times$ & \textbf{3.37} & \underline{4.493} & 3.95 & 4.498 & 5.97 & \underline{4.498} & \underline{6.43} & 4.485 & \underline{5.70} & 4.493\\
 \arrayrulecolor[gray]{0.5}
\cdashline{1-12}
\arrayrulecolor{black} 
\multirow{3}{*}{5} & $1\times$ & 4.14 & 4.485 & 4.48 & \underline{4.502} & 6.52 & 4.497 & 8.41 & \underline{4.491} & 6.84 & \underline{4.495} \\
\arrayrulecolor[gray]{0.5}
\cdashline{2-12}
\arrayrulecolor{black} 
 & $3\times$ & {3.43} & \textbf{4.495} & \textbf{3.65} & 4.498 & \underline{5.97} & 4.499 & \textbf{6.40} & 4.489 & \textbf{5.66} & \underline{4.495}\\
 \arrayrulecolor[gray]{0.5}
\cdashline{2-12}
\arrayrulecolor{black} 
 & $5\times$ & 3.84 & 4.478 & 4.28 & 4.493 & 6.40 & 4.489 & 7.70 & 4.483 & 6.46 & 4.486\\\arrayrulecolor[gray]{0.5}
\cdashline{1-12}
\arrayrulecolor{black} 
\multirow{4}{*}{7} & $1\times$ & 5.38 & 4.489 & 5.24 & \underline{4.502} & 7.59 & 4.480 & 9.23 & 4.490 & 7.79 & 4.487\\
\arrayrulecolor[gray]{0.5}
\cdashline{2-12}
\arrayrulecolor{black} 
 & $3\times$ & \underline{3.40} & 4.490 & \underline{3.92} & 4.499 & \textbf{5.91} & \underline{4.498} & {7.57} & \textbf{4.494} & 6.14 & \textbf{4.496}\\\arrayrulecolor[gray]{0.5}
\cdashline{2-12}
\arrayrulecolor{black} 
 & $5\times$ & 4.26 & 4.481 & 4.33 & 4.489 & 6.32 & 4.496 & 8.76 & 4.484 & 6.89 & 4.489 \\\arrayrulecolor[gray]{0.5}
\cdashline{2-10}
\arrayrulecolor{black} 
 & $7\times$ & 5.50 & 4.470 & 5.19 & 4.474 & 8.28 & 4.478 & 9.20 & 4.462 & 8.06 & 4.470\\\arrayrulecolor[gray]{0.5}
\cdashline{2-10}
\arrayrulecolor{black} 
\hline
\hline
  \end{tabular}
  }
\end{table}

For the other MTP implementations, the speedup ratio can be flexibly adjusted. In this study, we fix the number of MTP modules to 5 during training and evaluate performance at $3\times$ and $5\times$ speedup ratios during inference. For MTP-Parallel-Linear, the parallel linear layers disrupt the temporal dependencies between tokens, resulting in a noticeable drop in both WER and UTMOS with a higher speedup ratio. Similarly, for MTP-DeepSeek, performance degrades noticeably at the $5\times$ speedup ratio. This decline is likely due to the teacher-forcing next-step prediction strategy employed as noted in Section~\ref{sec:Implementation of MTP}. This approach does not enhance the model’s robustness against erroneous predictions, which becomes increasingly problematic as the speedup ratio rises. In contrast to previous methods, our proposed architecture demonstrates superior performance. Notably, even at a $5\times$ speedup ratio, the UTMOS remains high, and the WER remains exceptionally low. These results strongly validate the effectiveness of our MTP implementation, as it successfully addresses the issues in other methods.


\paragraph{Number of MTP Modules} To determine the optimal configuration for MTP modules, we conduct ablation studies on the number of MTP modules, as shown in Table~\ref{tab: result4_mpt_num}. The results indicate that the number used in the inference stage primarily affects modality alignment performance, with the best results typically achieved at a $3\times$ speedup ratio. Acoustic performance remains high, and only slightly decreases at higher speedup ratios. Overall, the number of MTP modules used during training has a relatively small impact, with the best performance achieved when training with 5 modules and infer at a $3\times$ speedup ratio. The results of VocalNet in Section~\ref{sec: overallresult} are also based on this configuration.

\begin{table}[htbp]
  \caption{Speech generation latency of VocalNet. Experiments are conducted on 1 NVIDIA L20 GPU.}
  \label{tab: latency_exp}
  \centering
  \resizebox{1\textwidth}{!}{
  \begin{tabular}{ccccccccc}
  \toprule
    \toprule
\textbf{Model} & \textbf{Speech Encoder (ms)} & \textbf{LLM (ms)} &\textbf{Speech Decoder (ms)} & \textbf{Speech Vocoder (ms)} & \textbf{Sum (ms)}  \\ 
\hline
\hline
VocalNet-1B & 35.86 & 33.95 & 24.74 & 225.18 & 319.73 \\
\arrayrulecolor[gray]{0.5}
\arrayrulecolor{black} 
VocalNet-8B & 36.08 & 126.71 & 40.02 & 225.56 & 428.38 \\\arrayrulecolor[gray]{0.5}
\arrayrulecolor{black} 
\hline
\hline
  \end{tabular}}
\end{table}



\subsection{Latency Analysis}


To provide a comprehensive evaluation of VocalNet, we perform a latency analysis, as presented in Table~\ref{tab: latency_exp}. The speech response delay is broken down into four distinct stages: first, the Whisper encoder processes the speech query; second, the LLM generates hidden states; third, the speech decoder predicts speech tokens; and finally, the speech vocoder constructs the response waveform. The latency calculations for the LLM and speech decoder are based on the decoding of 5 text tokens and 15 speech tokens, as described in Section~\ref{sec:model_config}, with a $3\times$ speedup ratio for the MTP decoder. The overall latency for VocalNet-1B and VocalNet-8B is approximately 320 ms and 430 ms, respectively. Notably, more than half of the latency is attributed to the speech vocoder, particularly during the flow-matching phase. These latency values were derived from tests conducted on a single L20 GPU.


\section{Conclusion}


In this paper, we present VocalNet-1B and VocalNet-8B, a series of advanced LLM-based speech interaction systems with high performance and low latency. We introduce multi-token prediction to accelerate speech token generation and enhance speech quality. Experiments on OpenAudioBench highlight the superior performance of VocalNet in voice assistant scenarios, showcasing its outstanding modality alignment and acoustic quality.

\medskip

{
\small

\bibliography{reference}

\begin{thebibliography}{35}
\providecommand{\natexlab}[1]{#1}
\providecommand{\url}[1]{\texttt{#1}}
\expandafter\ifx\csname urlstyle\endcsname\relax
  \providecommand{\doi}[1]{doi: #1}\else
  \providecommand{\doi}{doi: \begingroup \urlstyle{rm}\Url}\fi

\bibitem[An et~al.(2024)An, Chen, Deng, Du, Gao, Gao, Gu, He, Hu, Hu, et~al.]{an2024funaudiollm}
Keyu An, Qian Chen, Chong Deng, Zhihao Du, Changfeng Gao, Zhifu Gao, Yue Gu, Ting He, Hangrui Hu, Kai Hu, et~al.
\newblock Funaudiollm: Voice understanding and generation foundation models for natural interaction between humans and llms.
\newblock \emph{arXiv preprint arXiv:2407.04051}, 2024.

\bibitem[Berant et~al.(2013)Berant, Chou, Frostig, and Liang]{berant2013semantic}
Jonathan Berant, Andrew Chou, Roy Frostig, and Percy Liang.
\newblock Semantic parsing on freebase from question-answer pairs.
\newblock In \emph{Proceedings of the 2013 conference on empirical methods in natural language processing}, pages 1533--1544, 2013.

\bibitem[Cai et~al.(2024)Cai, Li, Geng, Peng, Lee, Chen, and Dao]{cai2024medusa}
Tianle Cai, Yuhong Li, Zhengyang Geng, Hongwu Peng, Jason~D Lee, Deming Chen, and Tri Dao.
\newblock Medusa: Simple llm inference acceleration framework with multiple decoding heads.
\newblock In \emph{International Conference on Machine Learning}, pages 5209--5235. PMLR, 2024.

\bibitem[Chen et~al.(2025)Chen, Chen, Chen, Chen, Chen, Deng, Du, Gao, Gao, Gao, et~al.]{chen2025minmo}
Qian Chen, Yafeng Chen, Yanni Chen, Mengzhe Chen, Yingda Chen, Chong Deng, Zhihao Du, Ruize Gao, Changfeng Gao, Zhifu Gao, et~al.
\newblock Minmo: A multimodal large language model for seamless voice interaction.
\newblock \emph{arXiv preprint arXiv:2501.06282}, 2025.

\bibitem[Chen et~al.(2024)Chen, Ma, Yan, Liang, Li, Xu, Niu, Zhu, Yang, Liu, et~al.]{chen2024slam}
Wenxi Chen, Ziyang Ma, Ruiqi Yan, Yuzhe Liang, Xiquan Li, Ruiyang Xu, Zhikang Niu, Yanqiao Zhu, Yifan Yang, Zhanxun Liu, et~al.
\newblock Slam-omni: Timbre-controllable voice interaction system with single-stage training.
\newblock \emph{arXiv preprint arXiv:2412.15649}, 2024.

\bibitem[D{\'e}fossez et~al.(2024)D{\'e}fossez, Mazar{\'e}, Orsini, Royer, P{\'e}rez, J{\'e}gou, Grave, and Zeghidour]{defossez2024moshi}
Alexandre D{\'e}fossez, Laurent Mazar{\'e}, Manu Orsini, Am{\'e}lie Royer, Patrick P{\'e}rez, Herv{\'e} J{\'e}gou, Edouard Grave, and Neil Zeghidour.
\newblock Moshi: a speech-text foundation model for real-time dialogue.
\newblock \emph{arXiv preprint arXiv:2410.00037}, 2024.

\bibitem[Du et~al.(2024)Du, Wang, Chen, Shi, Lv, Zhao, Gao, Yang, Gao, Wang, et~al.]{du2024cosyvoice}
Zhihao Du, Yuxuan Wang, Qian Chen, Xian Shi, Xiang Lv, Tianyu Zhao, Zhifu Gao, Yexin Yang, Changfeng Gao, Hui Wang, et~al.
\newblock Cosyvoice 2: Scalable streaming speech synthesis with large language models.
\newblock \emph{arXiv preprint arXiv:2412.10117}, 2024.

\bibitem[Fang et~al.(2024)Fang, Guo, Zhou, Ma, Zhang, and Feng]{fang2024llama}
Qingkai Fang, Shoutao Guo, Yan Zhou, Zhengrui Ma, Shaolei Zhang, and Yang Feng.
\newblock Llama-omni: Seamless speech interaction with large language models.
\newblock \emph{arXiv preprint arXiv:2409.06666}, 2024.

\bibitem[Fu et~al.(2025)Fu, Lin, Wang, Zhang, Shen, Liu, Li, Long, Gao, Li, et~al.]{fu2025vita}
Chaoyou Fu, Haojia Lin, Xiong Wang, Yi-Fan Zhang, Yunhang Shen, Xiaoyu Liu, Yangze Li, Zuwei Long, Heting Gao, Ke~Li, et~al.
\newblock Vita-1.5: Towards gpt-4o level real-time vision and speech interaction.
\newblock \emph{arXiv preprint arXiv:2501.01957}, 2025.

\bibitem[Gloeckle et~al.(2024)Gloeckle, Idrissi, Roziere, Lopez-Paz, and Synnaeve]{gloeckle2024better}
Fabian Gloeckle, Badr~Youbi Idrissi, Baptiste Roziere, David Lopez-Paz, and Gabriel Synnaeve.
\newblock Better \& faster large language models via multi-token prediction.
\newblock In \emph{International Conference on Machine Learning}, pages 15706--15734. PMLR, 2024.

\bibitem[Graves et~al.(2006)Graves, Fern{\'a}ndez, Gomez, and Schmidhuber]{graves2006connectionist}
Alex Graves, Santiago Fern{\'a}ndez, Faustino Gomez, and J{\"u}rgen Schmidhuber.
\newblock Connectionist temporal classification: labelling unsegmented sequence data with recurrent neural networks.
\newblock In \emph{Proceedings of the 23rd international conference on Machine learning}, pages 369--376, 2006.

\bibitem[Huang et~al.(2024)Huang, Li, Yang, Shi, Chang, Ye, Wu, Hong, Huang, Liu, et~al.]{huang2024audiogpt}
Rongjie Huang, Mingze Li, Dongchao Yang, Jiatong Shi, Xuankai Chang, Zhenhui Ye, Yuning Wu, Zhiqing Hong, Jiawei Huang, Jinglin Liu, et~al.
\newblock Audiogpt: Understanding and generating speech, music, sound, and talking head.
\newblock In \emph{Proceedings of the AAAI Conference on Artificial Intelligence}, volume~38, pages 23802--23804, 2024.

\bibitem[Ji et~al.(2024)Ji, Chen, Fang, Zuo, Lu, Wang, Jiang, Zhou, Liu, Cheng, et~al.]{ji2024wavchat}
Shengpeng Ji, Yifu Chen, Minghui Fang, Jialong Zuo, Jingyu Lu, Hanting Wang, Ziyue Jiang, Long Zhou, Shujie Liu, Xize Cheng, et~al.
\newblock Wavchat: A survey of spoken dialogue models.
\newblock \emph{arXiv preprint arXiv:2411.13577}, 2024.

\bibitem[Joshi et~al.(2017)Joshi, Choi, Weld, and Zettlemoyer]{joshi2017triviaqa}
Mandar Joshi, Eunsol Choi, Daniel~S Weld, and Luke Zettlemoyer.
\newblock Triviaqa: A large scale distantly supervised challenge dataset for reading comprehension.
\newblock In \emph{Proceedings of the 55th Annual Meeting of the Association for Computational Linguistics (Volume 1: Long Papers)}, pages 1601--1611, 2017.

\bibitem[Kong et~al.(2020)Kong, Kim, and Bae]{kong2020hifi}
Jungil Kong, Jaehyeon Kim, and Jaekyoung Bae.
\newblock Hifi-gan: Generative adversarial networks for efficient and high fidelity speech synthesis.
\newblock \emph{Advances in neural information processing systems}, 33:\penalty0 17022--17033, 2020.

\bibitem[Li et~al.(2025{\natexlab{a}})Li, Wang, Zhang, Guo, and Yu]{li2025fast}
Bohan Li, Hankun Wang, Situo Zhang, Yiwei Guo, and Kai Yu.
\newblock Fast and high-quality auto-regressive speech synthesis via speculative decoding.
\newblock In \emph{ICASSP 2025-2025 IEEE International Conference on Acoustics, Speech and Signal Processing (ICASSP)}, pages 1--5. IEEE, 2025{\natexlab{a}}.

\bibitem[Li et~al.(2023)Li, Zhang, Dubois, Taori, Gulrajani, Guestrin, Liang, and Hashimoto]{alpaca_eval}
Xuechen Li, Tianyi Zhang, Yann Dubois, Rohan Taori, Ishaan Gulrajani, Carlos Guestrin, Percy Liang, and Tatsunori~B. Hashimoto.
\newblock Alpacaeval: An automatic evaluator of instruction-following models.
\newblock \url{https://github.com/tatsu-lab/alpaca_eval}, 5 2023.

\bibitem[Li et~al.(2025{\natexlab{b}})Li, Liu, Zhang, Chen, Li, Li, Liu, Ming, Dong, Pan, et~al.]{li2025baichuan}
Yadong Li, Jun Liu, Tao Zhang, Song Chen, Tianpeng Li, Zehuan Li, Lijun Liu, Lingfeng Ming, Guosheng Dong, Da~Pan, et~al.
\newblock Baichuan-omni-1.5 technical report.
\newblock \emph{arXiv preprint arXiv:2501.15368}, 2025{\natexlab{b}}.

\bibitem[Li et~al.(2024)Li, Wei, Zhang, and Zhang]{li2024eagle}
Yuhui Li, Fangyun Wei, Chao Zhang, and Hongyang Zhang.
\newblock Eagle: speculative sampling requires rethinking feature uncertainty.
\newblock In \emph{Proceedings of the 41st International Conference on Machine Learning}, pages 28935--28948, 2024.

\bibitem[Liu et~al.(2024)Liu, Feng, Xue, Wang, Wu, Lu, Zhao, Deng, Zhang, Ruan, et~al.]{liu2024deepseek}
Aixin Liu, Bei Feng, Bing Xue, Bingxuan Wang, Bochao Wu, Chengda Lu, Chenggang Zhao, Chengqi Deng, Chenyu Zhang, Chong Ruan, et~al.
\newblock Deepseek-v3 technical report.
\newblock \emph{arXiv preprint arXiv:2412.19437}, 2024.

\bibitem[Mitsui et~al.(2024)Mitsui, Mitsuda, Wakatsuki, Hono, and Sawada]{mitsui2024pslm}
Kentaro Mitsui, Koh Mitsuda, Toshiaki Wakatsuki, Yukiya Hono, and Kei Sawada.
\newblock Pslm: Parallel generation of text and speech with llms for low-latency spoken dialogue systems.
\newblock \emph{arXiv preprint arXiv:2406.12428}, 2024.

\bibitem[Nachmani et~al.(2023)Nachmani, Levkovitch, Hirsch, Salazar, Asawaroengchai, Mariooryad, Rivlin, Skerry-Ryan, and Ramanovich]{nachmani2023spoken}
Eliya Nachmani, Alon Levkovitch, Roy Hirsch, Julian Salazar, Chulayuth Asawaroengchai, Soroosh Mariooryad, Ehud Rivlin, RJ~Skerry-Ryan, and Michelle~Tadmor Ramanovich.
\newblock Spoken question answering and speech continuation using spectrogram-powered llm.
\newblock \emph{arXiv preprint arXiv:2305.15255}, 2023.

\bibitem[Nguyen et~al.(2025)Nguyen, Muller, Yu, Costa-Jussa, Elbayad, Popuri, Ropers, Duquenne, Algayres, Mavlyutov, et~al.]{nguyen2025spirit}
Tu~Anh Nguyen, Benjamin Muller, Bokai Yu, Marta~R Costa-Jussa, Maha Elbayad, Sravya Popuri, Christophe Ropers, Paul-Ambroise Duquenne, Robin Algayres, Ruslan Mavlyutov, et~al.
\newblock Spirit-lm: Interleaved spoken and written language model.
\newblock \emph{Transactions of the Association for Computational Linguistics}, 13:\penalty0 30--52, 2025.

\bibitem[OpenAI(2024)]{gpt4o}
OpenAI.
\newblock \url{https://openai.com/index/hello-gpt-4o/}, 2024.

\bibitem[OpenBMB()]{MiniCPM-o-2.6}
OpenBMB.
\newblock Minicpm-o 2.6: A gpt-4o level mllm for vision, speech, and multimodal live streaming on your phone.
\newblock \url{https://openbmb.notion.site/185ede1b7a558042b5d5e45e6b237da9}.
\newblock Accessed: 2025-03-28.

\bibitem[Qi et~al.(2020)Qi, Yan, Gong, Liu, Duan, Chen, Zhang, and Zhou]{qi2020prophetnet}
Weizhen Qi, Yu~Yan, Yeyun Gong, Dayiheng Liu, Nan Duan, Jiusheng Chen, Ruofei Zhang, and Ming Zhou.
\newblock Prophetnet: Predicting future n-gram for sequence-to-sequencepre-training.
\newblock \emph{Findings of the Association for Computational Linguistics: EMNLP 2020}, 2020.

\bibitem[Radford et~al.(2023)Radford, Kim, Xu, Brockman, McLeavey, and Sutskever]{radford2023robust}
Alec Radford, Jong~Wook Kim, Tao Xu, Greg Brockman, Christine McLeavey, and Ilya Sutskever.
\newblock Robust speech recognition via large-scale weak supervision.
\newblock In \emph{International conference on machine learning}, pages 28492--28518. PMLR, 2023.

\bibitem[Saeki et~al.(2022)Saeki, Xin, Nakata, Koriyama, Takamichi, and Saruwatari]{saeki2022utmos}
Takaaki Saeki, Detai Xin, Wataru Nakata, Tomoki Koriyama, Shinnosuke Takamichi, and Hiroshi Saruwatari.
\newblock Utmos: Utokyo-sarulab system for voicemos challenge 2022.
\newblock \emph{arXiv preprint arXiv:2204.02152}, 2022.

\bibitem[Shen et~al.(2023)Shen, Song, Tan, Li, Lu, and Zhuang]{shen2023hugginggpt}
Yongliang Shen, Kaitao Song, Xu~Tan, Dongsheng Li, Weiming Lu, and Yueting Zhuang.
\newblock Hugginggpt: Solving ai tasks with chatgpt and its friends in hugging face.
\newblock \emph{Advances in Neural Information Processing Systems}, 36:\penalty0 38154--38180, 2023.

\bibitem[Wang et~al.(2024)Wang, Li, Fu, Shen, Xie, Li, Sun, and Ma]{wang2024freeze}
Xiong Wang, Yangze Li, Chaoyou Fu, Yunhang Shen, Lei Xie, Ke~Li, Xing Sun, and Long Ma.
\newblock Freeze-omni: A smart and low latency speech-to-speech dialogue model with frozen llm.
\newblock \emph{arXiv preprint arXiv:2411.00774}, 2024.

\bibitem[Xie and Wu(2024)]{xie2024mini}
Zhifei Xie and Changqiao Wu.
\newblock Mini-omni: Language models can hear, talk while thinking in streaming.
\newblock \emph{arXiv preprint arXiv:2408.16725}, 2024.

\bibitem[Xu et~al.(2025)Xu, Guo, He, Hu, He, Bai, Chen, Wang, Fan, Dang, Zhang, Wang, Chu, and Lin]{xu2025qwen}
Jin Xu, Zhifang Guo, Jinzheng He, Hangrui Hu, Ting He, Shuai Bai, Keqin Chen, Jialin Wang, Yang Fan, Kai Dang, Bin Zhang, Xiong Wang, Yunfei Chu, and Junyang Lin.
\newblock Qwen2.5-omni technical report.
\newblock \emph{arXiv preprint arXiv:2503.20215}, 2025.

\bibitem[Zeng et~al.(2024)Zeng, Du, Liu, Wang, Jiang, Zhao, Dong, and Tang]{zeng2024glm}
Aohan Zeng, Zhengxiao Du, Mingdao Liu, Kedong Wang, Shengmin Jiang, Lei Zhao, Yuxiao Dong, and Jie Tang.
\newblock Glm-4-voice: Towards intelligent and human-like end-to-end spoken chatbot.
\newblock \emph{arXiv preprint arXiv:2412.02612}, 2024.

\bibitem[Zhang et~al.(2024{\natexlab{a}})Zhang, Cheng, Deng, Chen, Wang, Zheng, Liu, Yu, Tan, Du, et~al.]{zhang2024omniflatten}
Qinglin Zhang, Luyao Cheng, Chong Deng, Qian Chen, Wen Wang, Siqi Zheng, Jiaqing Liu, Hai Yu, Chaohong Tan, Zhihao Du, et~al.
\newblock Omniflatten: An end-to-end gpt model for seamless voice conversation.
\newblock \emph{arXiv preprint arXiv:2410.17799}, 2024{\natexlab{a}}.

\bibitem[Zhang et~al.(2024{\natexlab{b}})Zhang, Lyu, Du, Chen, Zhang, Hu, Tan, Zhao, Wang, Zhang, et~al.]{zhang2024intrinsicvoice}
Xin Zhang, Xiang Lyu, Zhihao Du, Qian Chen, Dong Zhang, Hangrui Hu, Chaohong Tan, Tianyu Zhao, Yuxuan Wang, Bin Zhang, et~al.
\newblock Intrinsicvoice: Empowering llms with intrinsic real-time voice interaction abilities.
\newblock \emph{arXiv preprint arXiv:2410.08035}, 2024{\natexlab{b}}.

\end{thebibliography}


}


\appendix

\end{document}